\documentclass[runningheads]{llncs}
\usepackage{graphicx}
\usepackage{amsmath,amssymb} % define this before the line numbering.
\usepackage{color}
\usepackage{threeparttable}
\begin{document}

% \linenumbers

\title{``Factual'' or ``Emotional'': Stylized Image Captioning with Adaptive Learning and Attention} % Replace with your title
% Replace with your title

\titlerunning{Stylized Image Captioning with Adaptive Learning and Attention}
% Replace with a meaningful short version of your title

\authorrunning{T.Chen \and Z.Zhang \and Q.You \and C.Fang \and Z.Wang \and H.Jin \and J.Luo}
% Replace with shorter version of the author list. If there are more authors than fits a line, please use A. Author et al.

\author{Tianlang Chen\inst{1} \and Zhongping Zhang\inst{1} \and Quanzeng You\inst{3} \and \\ Chen Fang\inst{2} \and  Zhaowen Wang\inst{2} \and Hailin Jin\inst{2} \and Jiebo Luo\inst{1}}

%Please write out author names in full in the paper, i.e. full given and family names. 
%If any authors have names that can be parsed into FirstName LastName in multiple ways, please include the correct parsing, in a comment to the volume editors:
%\index{Lastnames, Firstnames}
%(Do not uncomment it, because you may introduce extra index items if you do that, we will use scripts for introducing index entries...)

\institute{
	University of Rochester,\email{ \{tchen45,jluo\}@cs.rochester.edu}, \email{ \{zzhang76\}@ur.rochester.edu}\and
    Adobe Research,\email{ \{cfang,zhawang,hljin\}@adobe.com} \and
    Microsoft Research,\email{ \{quyou\}@microsoft.com}
}

\maketitle

\begin{abstract}
Generating stylized captions for an image is an emerging topic in image captioning. Given an image as input, it requires the system to generate a caption that has a specific style (e.g., humorous, romantic, positive, and negative) while describing the image content semantically accurately. In this paper, we propose a novel stylized image captioning model that effectively takes both requirements into consideration. To this end, we first devise a new variant of LSTM, named style-factual LSTM, as the building block of our model. It  uses two groups of matrices to capture the factual and stylized knowledge, respectively, and automatically learns the word-level weights of the two groups based on previous context. In addition, when we train the model to capture stylized elements, we propose an adaptive learning approach based on a reference factual model, it provides factual knowledge to the model as the model learns from stylized caption labels, and can adaptively compute how much information to supply at each time step. We evaluate our model on two stylized image captioning datasets, which contain humorous/romantic captions and positive/negative captions, respectively. Experiments shows that our proposed model outperforms the state-of-the-art approaches, without using extra ground truth supervision. 
\end{abstract}

\section{Introduction}
\begin{figure}[!t]
\centering
\includegraphics[scale=0.33]{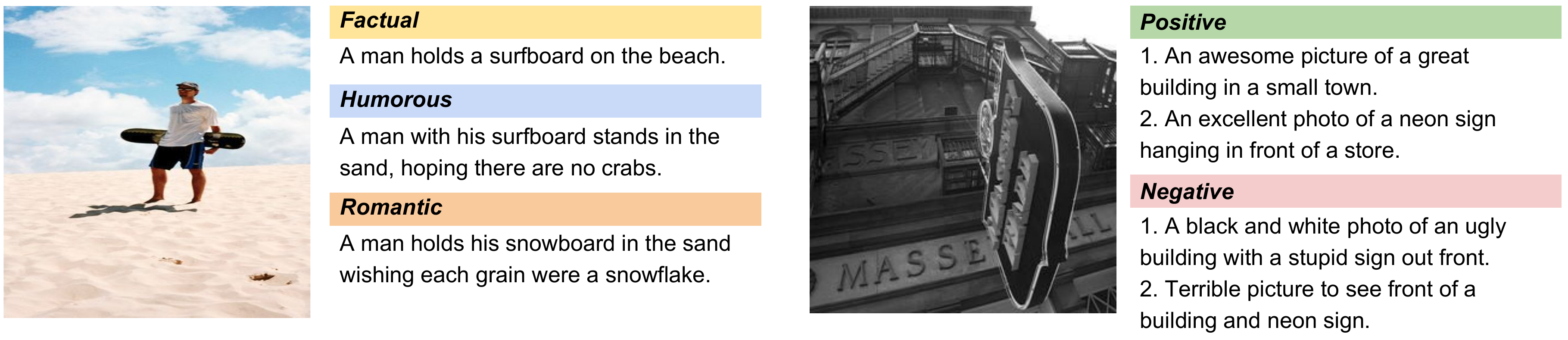}
 %where an .eps filename suffix will be assumed under latex,
 %and a .pdf suffix will be assumed for pdflatex; or what has been declared
 %via \DeclareGraphicsExtensions.

\caption{Examples of stylized image captions. Besides factual captions, there can be four kinds of stylized captions that correspond to humorous, romantic, positive and negative styles, respectively.}
\label{fig:overview}

\end{figure}

Automatically generating coherent captions for images has attracted remarkable attention for its strong applicability, such as picture auto-commenting \cite{li2016share} and helping blind people to see \cite{gurari2018vizwiz}. This task is often referred to as image captioning, which combines computer vision, natural language processing and artificial intelligence. Most recent image captioning systems focus on generating an objective, neutral and indicative caption without any style characteristics, which is defined as a factual caption. However, the art of language motivates researchers to generate captions with different styles, which can give people different feelings when focusing on a specific image. The ``style'' can refer to multiple meanings. For example, as shown in Figure~\ref{fig:overview}, in terms of the fashion of the caption, caption style can be either ``romantic'' or ``humorous''. In addition, in terms of the sentiment it brings to people, caption style can be either ``positive'' or ``negative''. Without doubt, generating such kinds of captions with different styles will greatly enrich the expressibility of the captions and make them more attractive.    

Ideally, a high-performing stylized image captioning model should satisfy two requirements: 1) it generates appropriate stylized words/phrases in appropriate positions of the caption, 2) it still describes the image content accurately. Focused on stylized caption generation, existing state-of-the-art work \cite{mathews2016senticap}\cite{gan2017stylenet} train their captioning models based on two datasets separately, a large dataset with paired images and ground truth factual captions, and a small dataset with paired images and stylized ground truth captions. From the large factual dataset, the model is learned to generate factual captions that can correctly describe the images; from the small stylized dataset, the model is learned to transform factual captions to stylized captions by incorporating suitable non-factual words/phrases at correct positions of the caption. In the training and predicting process, how to effectively take these two aspects into consideration is paramount for the model to generate high quality stylized captions.

To combine and preserve the knowledges learned from both factual and stylized dataset, Gan et al. \cite{gan2017stylenet} propose a factored LSTM, which factorizes matrix $W_{x\cdot}$ into three matrices $(U_{x\cdot}$, $S_{x\cdot}$, $V_{x\cdot})$. $U_{x\cdot}$ and $V_{x\cdot}$ are updated by the ground truth factual captions 
while $S_{x\cdot}$ is updated by ground truth captions with a specific style. In the predicting process, $U_{x\cdot}$, $S_{x\cdot}$ and $V_{x\cdot}$ are combined to generate the stylized caption. Since $U_{x\cdot}$ and $V_{x\cdot}$ preserve  the factual information and $S_{x\cdot}$ preserves the stylized information, the model can thus generate stylized captions that correspond to input images. However, for both the training and predicting processes, factored LSTM cannot differentiate whether paying more attention to the fact-related part (i.e. $U_{x\cdot}$ and $V_{x\cdot}$) or the style-related part (i.e. $S_{x\cdot}$). It is natural that when the model focuses on predicting a stylized word, it should pay more attention to the style-related part, and vice versa. Mathews et al. \cite{mathews2016senticap} consider this problem and propose Senticap, which consists of two parallel LSTMs -- one updated by the factual captions and one updated by the sentimental captions. When predicting a word, Senticap obtains the result by weighting the predicted word probability distributions of the two LSTMs. However, directly weighting the high level probability distributions can be too ``coarse'' in that it doesn't consider the low level attention effect on stylized and factual elements. In addition, Senticap obtains the weights of the two distributions by predicting the sentiment strength of the current word. In this step, it uses the extra ground truth word sentiment strength label, which is unavailable for other datasets.

In this paper, we propose a novel stylized image captioning model. In particular, we first design a new style-factual LSTM as a core building block of our model. Compared with factored LSTM, it combines fact-related and style-related parts of LSTM in a different way and incorporates self-attention for this two parts. More concretely, for both input word embedding feature and input hidden state of LSTM, we assign two independent groups of matrices to capture the factual and stylized knowledges, respectively. At each time step, it feeds an effective attention mechanism to weight the importance of the two groups of parameters based on previous context information, and combines the two groups of parameters by weighted-sum operation. In addition, to help the model preserve factual information while learning from stylized captions, we develop an adaptive learning approach that feeds a reference factual model as a guidance. At each time step, the model can adaptively learn whether to focus more on the ground truth stylized label or on the factual guidance, based on the similarity between the outputs of the real stylized captioning model and the reference factual model. Overall, both improvements help the model capture and combine the factual and stylized knowledge in a better way.      

In summary, the main contributions of this paper are:
\begin{list}{$\bullet$}
{ \setlength{\leftmargin}{0.5em}}
    \item We propose a new stylized image captioning model, with a core building block named style-factual LSTM. Style-factual LSTM incorporates two groups of parameters with dynamic attention weights into an LSTM, to adaptively adjust the relative attention weights between the fact and style-related parts.
    \item We develop a new learning approach to training the model on stylized captions, which adds the factual output of reference model as a guidance. The model can automatically adjust the strength of guidance based on ground truth stylized caption and reference model output without using additional information.       
    \item Our model outperforms the state-of the-art methods on both image style captioning and image sentiment captioning task, in terms of both the relevance to the image and the appropriateness of the style.
    \item We visualize the corresponding attention weights for both the style-factual LSTM and the adaptive learning approach, and show explainable improvements in the results.  
\end{list}

\section{Related Work}
Stylized image captioning is mainly related to two research topics: image captioning and style transfer. In this section, we provide the background of image captioning, attention model and style transfer, respectively.

\textit{Image Captioning.}
Image captioning has received much attention in recent years due to the advances in computer vision and natural language processing.
Early image captioning methods \cite{farhadi2010every}\cite{fang2015captions}\cite{kulkarni2011baby}\cite{li2011composing}\cite{kuznetsova2012collective}\cite{lebret2014simple}\cite{elliott2013image} generate sentences by combining words which are extracted from corresponding images. A downside of these methods is that their performance is limited by empirical language models. To alleviate the problem, retrieval-based frameworks are developed \cite{kuznetsova2014treetalk}\cite{ordonez2011im2text}\cite{hodosh2013framing}\cite{kuznetsova2012collective}. They first retrieve similarity images of the input image from a database, then generate new descriptions for the query image by using the captions of retrieved images. However, this kind of approach relies heavily on the image database. Modern approaches \cite{karpathy2015deep}\cite{donahue2015long}\cite{mao2015learning}\cite{chen2015mind}\cite{mao2014deep}\cite{vinyals2015show}\cite{xu2015show}\cite{tran2016rich} consider image captioning as a machine translation problem. Vinyals et al. \cite{vinyals2015show} propose an encoder-decoder framework. Many improved approaches \cite{karpathy2015deep}\cite{donahue2015long}\cite{mao2015learning}\cite{mnih2014recurrent}\cite{xu2015show}\cite{tran2016rich} are developed based on this encoder-decoder framework. The differences between these methods often lie in the architecture  of recurrent neural network.

\textit{Attention Model.} 
Recent successes of attention models \cite{sutskever2014sequence}\cite{rush2015neural}\cite{hermann2015teaching}\cite{rocktaschel2015reasoning}\cite{bahdanau2014neural} motivate many researchers to apply visual or language attention models \cite{xu2015show}\cite{mnih2014recurrent} \cite{lu2017knowing}\cite{spratling2004feedback}\cite{you2016image}\cite{anderson2017bottom} to the image captioning task.  Top-down visual attention models are first widely used \cite{mnih2014recurrent}\cite{wu2016encode}\cite{xu2015show}\cite{tang2014learning}. The attention models enable deeper image understanding by assigning different attention weights to different image regions. Bottom-up and top-down combined attention models \cite{you2016image}\cite{anderson2017bottom} are also proposed to take even one step further. In \cite{lu2017knowing}, the authors propose a novel adaptive attention model with a visual sentinel. This model not only can determine where to attend to in images, but also adaptively decide whether it needs to attend the image or to the LSTM decoder according to different words. Motivated by this work, we develop a novel joint style-factual attention architecture to make the model adaptively learns from the factual part and stylized part.

\textit{Style Transfer.} Most style transfer works \cite{gatys2016image}\cite{johnson2016perceptual}\cite{neumann2005color}\cite{ulyanov2016texture} focus on image style transfer. These works utilize the Gram matrix of hidden layers to measure the distance between different styles. In the meantime, pure text style transfer is making breakthrough as the development of nature language processing. For example, Shen et al. \cite{shen2017style} propose a cross-alignment method to transfer text into different styles by generating a shared latent content space. Hu et al. \cite{hu2017toward} propose a neural generative model that combines variational auto-encoders (VAEs) and holistic attribute discriminators, to generate sentences while controlling the attributes. Combined with the above topics, in recent years, researchers begin to focus on stylized image captioning. Gan et al. and Mathews et al. propose StyleNet \cite{gan2017stylenet} and SentiCap \cite{mathews2016senticap} to generate image captions with specific styles and sentiments, respectively.  Along the same direction, we propose a novel stylized image captioning model that achieves promising performance on both tasks.

\section{Method}\label{sec:app}
In this section, we formally present our stylized image  captioning model. Specifically, we introduce the basic encoder-decoder image captioning model in Section~\ref{sec:basic}. In Section~\ref{sec:sfLSTM}, we present style-factual LSTM as the core building block of our framework. In Section~\ref{sec:learn}, we present the overall learning strategy of the style-factual LSTM and in Section~\ref{sec:refer}, we describe an adaptive learning approach to help the model generate stylized captions without deviating from the related image content.

\subsection{Encoder-decoder Image Captioning Model}\label{sec:basic}

We first describe the basic encoder-decoder model \cite{vinyals2015show} for image caption generation. Giving an image $I$ and its corresponding caption \textbf{y} = $\{y_{1}, ... ,y_{T}\}$, the encoder-decoder model minimizes the following maximum likelihood estimation (MLE) loss function:

\begin{equation}
\begin{aligned}
\theta^{*} = \mathop{\arg\min}_{\theta}\sum_{I,\textbf{y}}\log{p(\textbf{y}|I;\theta)}
\end{aligned}
\end{equation}

\noindent where $\theta$ denotes the parameters of the model. By applying chain rule, the log likelihood of the joint probability distribution can be expressed as follows:

\begin{equation}
\begin{aligned}
\log{p(\textbf{y})} = \sum_{t = 1}^{T}\log{p(y_{t}|y_{1}, ... ,y_{t-1},I)}
\end{aligned}
\end{equation}

\noindent where we drop the dependency on $\theta$ for convenience. 

For the encoder-decoder image captioning model, LSTM is commonly used to model $p(y_{t}|y_{1}, ... ,y_{t-1},I)$. Specifically, it can be expressed as:

\begin{equation}
\begin{aligned}
p(y_{t+1}|y_{1}, ... ,y_{t},I) = f(h_{t}) \\
h_{t} = LSTM(x_{t},h_{t-1})
\end{aligned}
\end{equation}

\noindent where $h_{t}$ is the hidden state of LSTM at time $t$, $f(\cdot)$ is a non-linear sub-network which maps $h_{t}$ into word probability distribution. For $t > 0$, $x_{t}$ is the word embedding feature of word $y_{t}$; for $t = 0$, $x_{0}$ is the image feature of $I$.

\subsection{Style-factual LSTM}\label{sec:sfLSTM}

\begin{figure}[!t]
\centering
\includegraphics[width=3.1in]{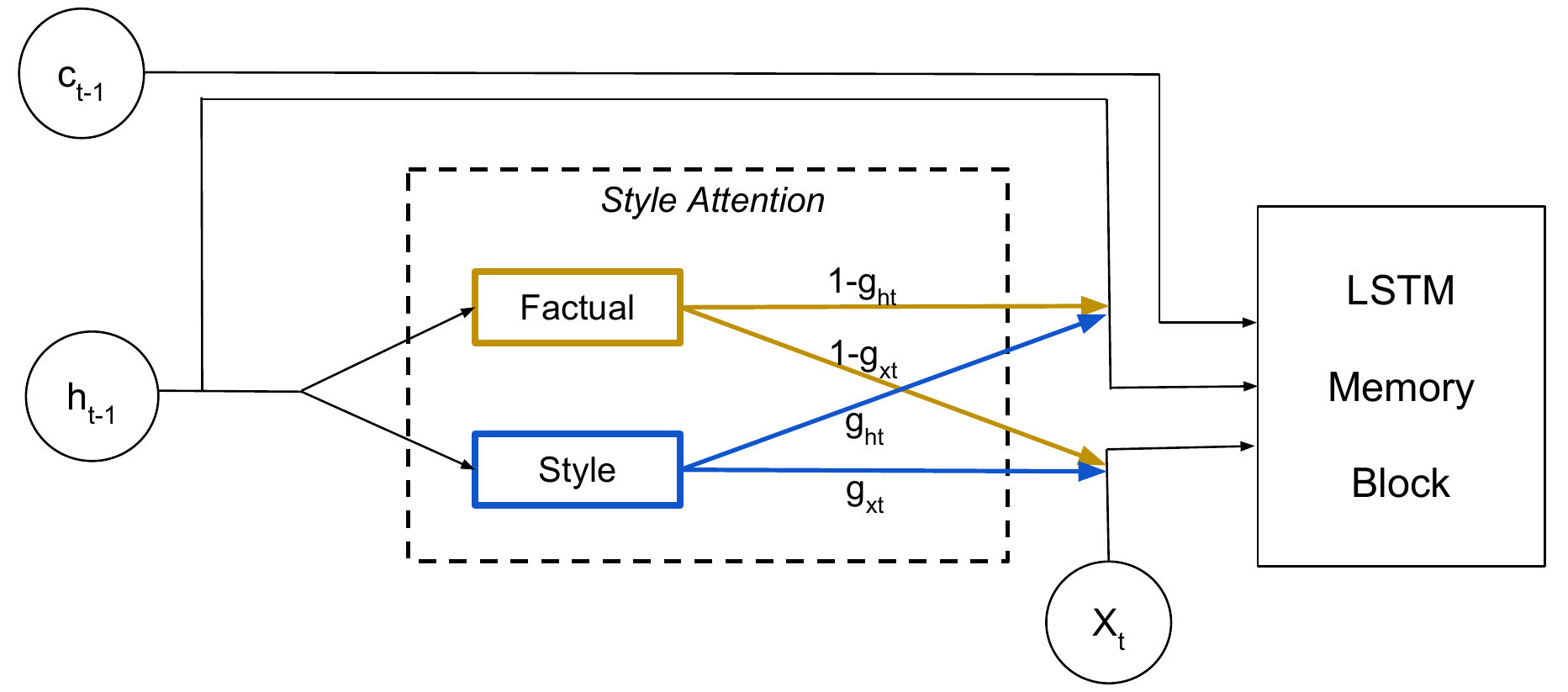}
 %where an .eps filename suffix will be assumed under latex,
 %and a .pdf suffix will be assumed for pdflatex; or what has been declared
 %via \DeclareGraphicsExtensions.
\caption{An illustration of the style-factual LSTM block. Four weights, $1-g_{ht}$, $1-g_{xt}$, $g_{ht}$ and $g_{xt}$, are designed to control the proportions of $W_{hi}$, $W_{xi}$, $S_{hi}$ and $S_{xi}$  matrices, respectively.}
\label{fig:style_attention}
\end{figure}

To make our model capable of generating a stylized caption consistent with the image content, we devise the style-factual LSTM, which feeds two new groups of matrices $S_{x\cdot}$ and $S_{h\cdot}$ as the counterparts of $W_{x\cdot}$ and $W_{h\cdot}$, to learn to stylize the caption. In addition, at time step $t$, adaptive weights $g_{xt}$ and $g_{ht}$ are synchronously learned to adjust the relative attention weights between $W_{x\cdot}$ and $S_{x\cdot}$ as well as $W_{h\cdot}$ and $S_{h\cdot}$. The structure of style-factual LSTM is shown as Figure~\ref{fig:style_attention}. In particular, the style-factual LSTM are defined as follows:
\begin{equation}\label{equ:lstm}
\begin{aligned}
i_t &= \sigma((g_{xt}S_{xi} + (1-g_{xt})W_{xi})x_t+(g_{ht}S_{hi} + (1-g_{ht})W_{hi})h_{t-1}+b_i) \\
f_t &= \sigma((g_{xt}S_{xf} + (1-g_{xt})W_{xf})x_t+(g_{ht}S_{hf} + (1-g_{ht})W_{hf})h_{t-1}+b_f) \\
o_t &= \sigma((g_{xt}S_{xo} + (1-g_{xt})W_{xo})x_t+(g_{ht}S_{ho} + (1-g_{ht})W_{ho})h_{t-1}+b_o) \\
\widetilde{c}_t &= \phi((g_{xt}S_{xc} + (1-g_{xt})W_{xc})x_t+(g_{ht}S_{hc} + (1-g_{ht})W_{hc})h_{t-1}+b_c) \\
c_t &= f_t\odot c_{t-1}+i_t\odot \widetilde{c}_t \\
h_t &= o_t\odot\phi(c_t)
\end{aligned}
\end{equation}
\noindent where $W_{x\cdot}$ and $W_{h\cdot}$ are responsible for generating the factual caption based on the input image, while $S_{x\cdot}$ and $S_{h\cdot}$ are responsible for adding specific style into the caption. At time step $t$, the style-factual LSTM feeds $h_{t-1}$ into two independent sub-networks with one output node, which in the end figures out $g_{xt}$ and $g_{ht}$ after using the $sigmoid$ unit to map the outputs to the range of (0, 1). Intuitively, when the model aims to predict a factual word, $g_{xt}$ and $g_{ht}$ should be close to 0, which encourages the model to predict the word based on $W_{x\cdot}$ and $W_{h\cdot}$. On the other hand, when the model focuses on predicting a stylized word, $g_{xt}$ and $g_{ht}$ should be close to 1, which encourages the model to predict the word based on $S_{x\cdot}$ and $S_{h\cdot}$. 

\subsection{Overall Learning Strategy}\label{sec:learn}
Similar to \cite{gan2017stylenet}\cite{luong2015multi}, we adopt a two-stage learning strategy to train our model. For each epoch, our model is sequentially trained by two independent stages. In the first stage, we manually fix $g_{xt}$ and $g_{ht}$ to 0, freezing the style-related matrices $S_{x\cdot}$ and $S_{h\cdot}$. We train the model using the paired images and ground truth factual captions. In accordance with \cite{vinyals2015show}, for an image-caption pair, we first extract the deep-level feature of the image using a pre-trained CNN, and then map it into an appropriate space by a linear transformation matrix. For each word, we embed its corresponding one-hot vector by a word embedding layer such that each word embedding feature has the same dimension as the transformed image feature. During training, the image feature is only fed into the LSTM as an input at the first time step. In this stage, for the style-factual LSTM, only $W_{x\cdot}$ and $W_{h\cdot}$ are updated with other layers' parameters so that they focus on generating factual captions without styles. As mentioned in Section~\ref{sec:basic}, the MLE loss is used to train the model.

In the second stage, $g_{xt}$ and $g_{ht}$ are learned by the two attention sub-networks mentioned in Section~\ref{sec:sfLSTM}, as this activates $S_{x\cdot}$ and $S_{h\cdot}$ to participate in generating the stylized caption. For this stage, we use the paired images and ground truth stylized captions to train our model. In particular, different from the first stage, we update $S_{x\cdot}$ and $S_{h\cdot}$ for style-factual LSTM, with $W_{x\cdot}$ and $W_{h\cdot}$ fixed. Also, the parameters of the two attention sub-networks are updated concurrently with the whole network. Instead of only using the MLE loss, in Section~\ref{sec:refer}, we will propose a novel approach to training our model in this stage.

For the test stage, to generate a stylized caption based on an image, we still compute $g_{xt}$ and $g_{ht}$ by the attention sub-networks, which activates $S_{x\cdot}$ and $S_{h\cdot}$. The classical beam search approach is used to predict the caption.

\subsection{Adaptive Learning with Reference Factual Model}\label{sec:refer}
Our goal is to generate stylized captions that can accurately describe the image at the same time. Considering our style-factual LSTM, if we directly use the MLE loss to update $S_{x\cdot}$ and $S_{h\cdot}$ based on Section~\ref{sec:learn}, it will only be updated via a few ground truth stylized captions, without learning anything from the much more massive ground truth factual captions. This may lead to the situation where the generated stylized caption cannot describe the images well. Intuitively, in a specific time step, when the generated word is unrelated to style, we encourage the model to learn more from the ground truth factual captions, instead of just a small number of the ground truth stylized captions. %When the model focuses on predicting a factual word, we are willing to let it learn more from ground truth factual information, instead of just a small number of the ground truth stylized captions. 

\begin{figure}[!t]
\centering
\includegraphics[width=4.7in]{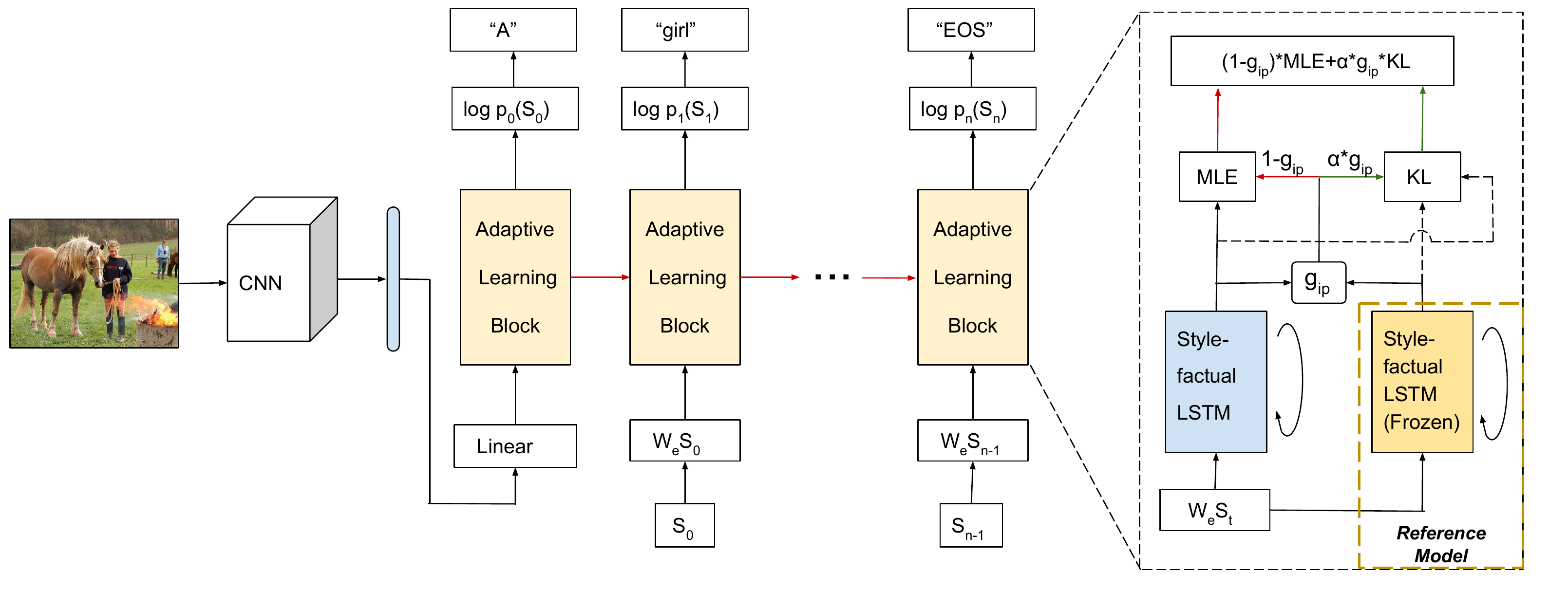}
 %where an .eps filename suffix will be assumed under latex,
 %and a .pdf suffix will be assumed for pdflatex; or what has been declared
 %via \DeclareGraphicsExtensions.
\caption{The framework of our stylized image captioning model. In the adaptive learning block, the style-related matrices in the reference model (yellow) are frozen. It is designed to lead the real style-factual LSTM (blue) to learn from factual information selectively.}

%Visual features of CNN responses  and attribute detections  are injected into RNN %%%dashed arrows and get fused together through a feedback loop (blue arrows). Attention %on attributes is enforced by both input model and output model.
%}
\label{fig:reference_factual}
\end{figure}

Motivated by this consideration, we propose an adaptive learning approach, for which the model concurrently learns information from the ground truth stylized captions and the reference factual model, and adaptively adjusts their relative learning strengths. 

In the second stage of the training process, giving an image and the corresponding ground truth stylized caption, in addition to predicting the stylized caption by the real model as Section~\ref{sec:learn}, the framework also gives the predicted ``factual version'' output based on the reference model. Specifically, for reference model, we set $g_{xt}$ and $g_{ht}$ to 0, which freezes $S_{x\cdot}$ and $S_{h\cdot}$ as the first training stage, so that the reference model will generate its output based on $W_{x\cdot}$ and $W_{h\cdot}$. Noted that $W_{x\cdot}$ and $W_{h\cdot}$ are trained by the ground truth factual captions. At time step $t$, denote the predicted word probability distribution by the real model as $P_{s}^{t}$, and the predicted word probability distribution by the reference model as $P_{r}^{t}$, we first compute their Kullback–Leibler divergence (KL-divergence) as follows:

\begin{equation}
\begin{aligned}
D(P_{s}^{t}||P_{r}^{t}) = \sum_{w \in W}P_{s}^{t}(w)\log{\frac{P_{s}^{t}(w)}{P_{r}^{t}(w)}}
\end{aligned}
\end{equation}

\noindent where $W$ is the word vocabulary. Intuitively, if the model focuses on generating a factual word, we aim to decrease $D(P_{s}^{t}||P_{r}^{t})$, which makes $P_{s}^{t}$ similar to $P_{r}^{t}$. In contrast, if the model focuses on generating a stylized word, we update the model by the MLE loss based on the corresponding  ground truth stylized word. 

To judge whether the current predicted word is related to style or not, we compute the inner product of $P_{s}^{t}$ and $P_{r}^{t}$ as the factual strength of the predicted word, we denote it as $g_{ip}^{t}$, and use it to adjust the weight between MLE and KL-divergence losses. In essence, $g_{ip}^{t}$ represents the similarity between the word probability distributions $P_{s}^{t}$ and $P_{r}^{t}$. When $g_{ip}^{t}$ is close to 0, $P_{s}^{t}$ has a higher possibility to correspond to a stylized word, because the reference model does not have the capacity to generate stylized words, which in the end makes $g_{ip}^{t}$ small. In this situation, a higher attention weight should be given to the MLE loss. On the other hand, when $g_{ip}^{t}$ is large, $P_{s}^{t}$ has a  higher possibility to correspond to a factual word, we then give KL-divergence losses higher significance.

The complete framework with the proposed adaptive learning approach is shown in Figure~\ref{fig:reference_factual}. In the end, the new loss function for the second training stage is expressed as follows:

\begin{equation}
\begin{aligned}
Loss = \sum_{t=1}^{T}-(1-g_{ip}^{t})logP_{s}^{t}(y_{t}) + \alpha \cdot \sum_{t=1}^{T} g_{ip}^{t}D(P_{s}^{t}||P_{r}^{t})
\end{aligned}
\end{equation}

\noindent where $\alpha$ is a hyper-parameter to control the relative importance of the two loss terms. In the training process, $g_{ip}^{t}$ and $P_{r}^{t}$ do not participate in the back propagation. Still, for the style-factual LSTM, only $S_{x\cdot}$, $S_{h\cdot}$ and parameters of two attention sub-networks are updated.

\section{Experiments}
We perform extensive experiments to evaluate the proposed models. Experiments are evaluated by standard image captioning measurements -- BLEU, Meteor,
Rouge-L and CIDEr. We will first discuss the
datasets and model settings used in the experiments. We then  compare and analyze the results of the proposed model with the state-of-the-art stylized image captioning models.
\subsection{Datasets and Model Settings}
At present, there are two datasets related to stylized image  captioning. First, Gan et al. \cite{gan2017stylenet} collect a FlickrStyle10K dataset that contains 10K Flickr images with stylized captions. It should be noted that only the 7K training set are public. In particular, for the 7K images, each image is labeled with 5 factual captions, 1 humorous caption and 1 romantic caption. We randomly select 6000 of them as the training set, and 1000 of them as the test set. For the training set, we randomly split 10\% of them as the validation set to adjust the hyper-parameters. Second, Mathews et al. \cite{mathews2016senticap} provide an image sentiment captioning dataset based on MSCOCO images, which contains images that are labeled by positive and negative sentiment captions. The POS subset contains 2,873 positive captions and 998 images for training, and another 2,019 captions over 673 images for testing. The NEG subset contains 2,468 negative captions and 997 images for training, and another 1,509 captions over 503 images for testing. Each of the test images has three positive and/or three negative captions. Following \cite{mathews2016senticap}, on the training process, this sentiment dataset can be used with MSCOCO training set \cite{chen2015microsoft} of 413K+ factual sentences on 82K+ images, as the factual training set. 

We extract image features by CNN. To make fair comparisons, for image sentiment captioning, we extract the 4096-dimension image feature by the second to last fully-connected layer of VGG-16 \cite{simonyan2014very}. For stylized image captioning, we extract the 2048-dimension image feature by the last pooling layer of ResNet152 \cite{he2016deep}. These settings are consistent with the corresponding works. Same as \cite{mathews2016senticap}, we set the dimension of both word embedding feature and LSTM hidden state to 512 (this setting applys to all the proposed and baseline models in our experiments). For both style captioning and sentiment captioning, we use the Adam algorithm for model updating with a mini-batch size of 64 for both stages. The learning rate is set to 0.001. For style captioning, the hyper-parameter $\alpha$ mentioned in Section~\ref{sec:refer} is set to 1.1, for sentiment captioning, $\alpha$ is set to 0.9 and 1.5 for positive and negative captioning, which leads to the best performance in the validation set. Also, for style captioning, we directly input images into ResNet without normalization, which achieves better performance.

\subsection{Performance on stylized image captioning Dataset}
\subsubsection{Experiment settings}
We first evaluate our proposed model on the style captioning dataset. Consistent with \cite{gan2017stylenet}, following baselines are used for comparison:

\begin{list}{$\bullet$}
{ \setlength{\leftmargin}{0.5em}}
    \item CaptionBot \cite{tran2016rich}: the commercial image captioning system released by Microsoft, which is trained on the large-scale factual image-caption pair data.  
    \item Neural Image Caption (NIC) \cite{vinyals2015show}: the standard encoder-decoder model for image captioning. It is trained by factual image-caption pairs of the training dataset and can generate factual captions.
    
    \item Fine-tuned: we first train an NIC, and then use the additional stylized image-caption pairs to update the parameters of the LSTM language model.
    \item StyleNet \cite{gan2017stylenet}: we train a StyleNet as \cite{gan2017stylenet}. To make fair comparisons, different from the original model that only uses stylized captions to update the parameter in the second stage, we train the model by the complete stylized image-caption pairs. It has two parallel model StyleNet(H) and StyleNet(R), which generate humorous and romantic captions, respectively .    
\end{list}
Our goal is to generate captions that are both appropriately stylized and consistent with the image. There are no definite ways to separately measure these two aspects. To measure them comprehensively, for stylized captions generated by different models, we compute the BLEU-1,2,3,4, ROUGE, CIDEr, METEOR scores based on both the ground truth stylized captions and ground truth factual captions. High-performance on both situations will demonstrate the effectiveness of the stylized image captioning model for both requirements. Because we split the dataset in a different way, we re-implement all the models and compute the scores instead of directly citing them from \cite{gan2017stylenet}.

\subsubsection{Experiment results}

\begin{table*}[htbp]
  \small\centering
  \caption{\label{tab:result1} BLEU-1,2,3,4, ROUGE, CIDEr, METEOR scores of the proposed model and state-of-the-art methods based on ground truth stylized and factual references. ``SF-LSTM'' and ``Adap'' represents style-factual LSTM and adaptive learning approach.}
  \begin{threeparttable}
  \scalebox{0.86}{
  \begin{tabular}{|c|c|c|c|c|c|c|c|}

    \cline{1-8}
    Model&BLEU-1&BLEU-2&BLEU-3&BLEU-4&ROUGE&CIDEr&METEOR\\ \cline{1-8}
    \multicolumn{8}{|c|}{Humorous/Factual Generations + Humorous References} \\ \cline{1-8}
    
    CaptionBot &19.7&9.5&5.1&2.8&22.8&28.1&8.9\\ 
    NIC &25.4&13.3&7.4&4.2&24.3&34.1&10.4\\ 
    Fine-tuned(H) &26.5&13.6&7.6&4.3&24.4&35.4&10.6\\ 
    StyleNet(H) &24.1&11.7&6.5&3.9&22.3&30.7&9.4\\ \cline{1-8}
    SF-LSTM(H) (ours)&26.8&14.2&8.2&4.9&24.8&\textbf{39.8}&\textbf{11.0}\\ 
    SF-LSTM + Adap(H) (ours)&\textbf{27.4}&\textbf{14.6}&\textbf{8.5}&\textbf{5.1}&\textbf{25.3}&39.5&\textbf{11.0}\\ \cline{1-8}
    
\cline{1-8}
    \multicolumn{8}{|c|}{Romantic/Factual Generations + Romantic References} \\ \cline{1-8}
    %Model&BLEU-1&BLEU-2&BLEU-3&BLEU-4&ROUGE&CIEDr&METEOR\\ \cline{1-8}
    CaptionBot &18.4&8.7&4.5&2.4&22.3&25.0&8.7\\ 
    NIC &24.3&12.8&7.4&4.4&24.1&33.7&10.2\\ 
    Fine-tuned(R) &26.8&13.6&7.7&4.6&24.8&36.6&11.0\\ 
    StyleNet(R) &25.4&11.7&6.1&3.5&23.2&27.9&10.0\\ \cline{1-8}
    SF-LSTM(R) (ours)&27.4&14.2&8.1&\textbf{4.9}&25.0&37.4&11.1\\ 
    SF-LSTM + Adap(R) (ours)&\textbf{27.8}&\textbf{14.4}&\textbf{8.2}&4.8&\textbf{25.5}&\textbf{37.5}&\textbf{11.2}\\ \cline{1-8}
    
\cline{1-8}

    \multicolumn{8}{|c|}{Humorous Generations + Factual References} \\ \cline{1-8}
    %Model&BLEU-1&BLEU-2&BLEU-3&BLEU-4&ROUGE&CIEDr&METEOR\\ \cline{1-8}
    Fine-tuned(H) &48.0&31.1&19.9&12.6&39.5&26.2&18.1\\ 
    StyleNet(H) &45.8&28.5&17.6&11.3&36.3&22.7&16.3\\ \cline{1-8}
    SF-LSTM(H) (ours)&47.8&31.7&20.6&13.1&39.8&28.2&18.7\\ 
    SF-LSTM + Adap(H) (ours)&\textbf{51.5}&\textbf{34.6}&\textbf{23.1}&\textbf{15.4}&\textbf{41.7}&\textbf{34.2}&\textbf{19.3}\\ \cline{1-8}
    
\cline{1-8}
    \multicolumn{8}{|c|}{Romantic Generations + Factual References} \\ \cline{1-8}
    %Model&BLEU-1&BLEU-2&BLEU-3&BLEU-4&ROUGE&CIEDr&METEOR\\ \cline{1-8}
    Fine-tuned(R) &46.4&30.4&20.2&13.5&38.5&24.0&18.2\\ 
    StyleNet(R) &44.2&26.8&16.3&10.4&35.4&15.8&16.3\\ \cline{1-8}
    SF-LSTM(R) (ours)&47.1&30.5&19.8&12.8&38.8&23.5&18.4\\ 
    SF-LSTM + Adap(R) (ours)&\textbf{48.2}&\textbf{31.5}&\textbf{20.6}&\textbf{13.5}&\textbf{40.2}&\textbf{26.7}&\textbf{18.7}\\ \cline{1-8}

    \end{tabular}}%
  \end{threeparttable}

\end{table*}%

Table~\ref{tab:result1} shows the quantitative results of different models based on different types of ground truth captions. Considering that for each image of the test set, we only have one ground truth stylized caption instead of five, excepts CIDEr, the overall performance of other measures based on the ground truth stylized captions is reasonably lower than \cite{gan2017stylenet}, because these measures are sensitive to the number of ground truth captions of each image. From the results, we can see that our proposed model achieves the best performance by almost all measures, regardless of testing on stylized or factual references. This demonstrates the effectiveness of our proposed model. In addition, we could see that feeding adaptive learning approach into our model can remarkably improve the scores based on factual references, for both humorous and romantic caption generations. This indicates the improvement for generated captions' affinity toward the images. Compared with directly training the model by stylized references using MLE loss, adaptive learning can guide the model to preserve factual information in a better way, when it focuses on generating a non-stylized word. 

\begin{figure*}[!htb]
\centering
\includegraphics[scale=0.28]{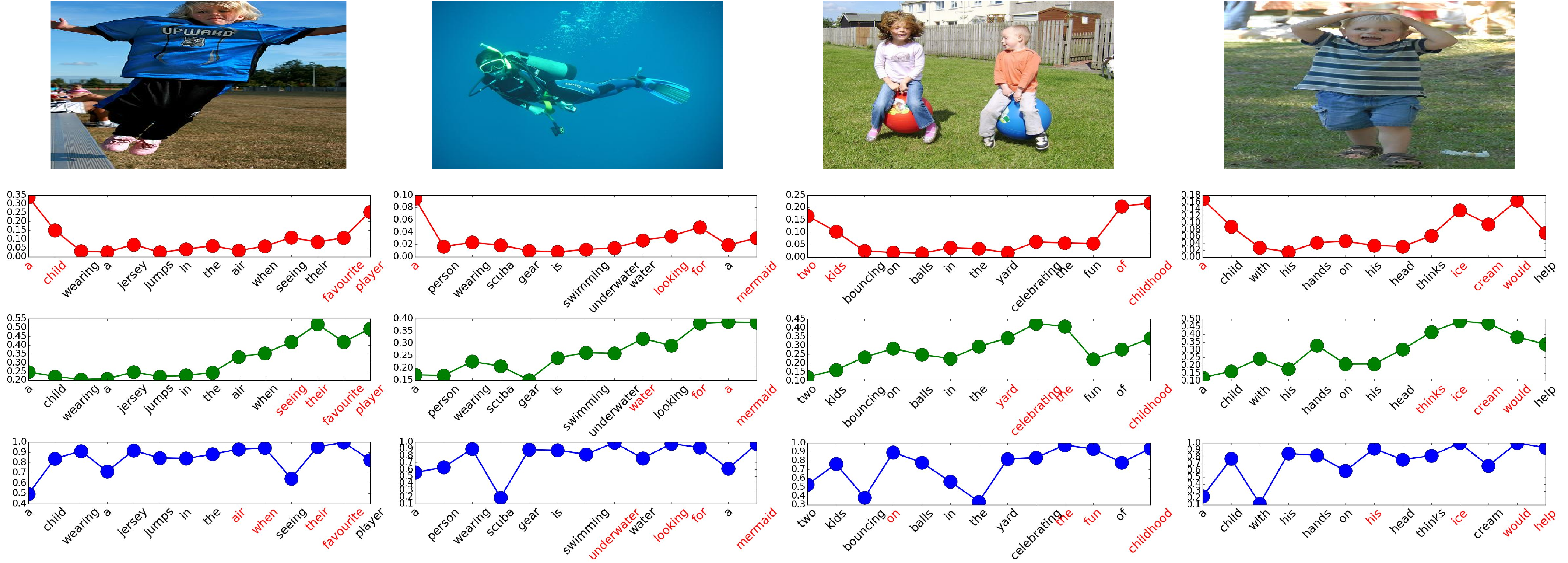}

\caption{Visualization of $g_{xt}$, $g_{ht}$ and $1 - g_{ip}$ on several examples. The \textit{second}, \textit{third} and \textit{fourth} rows correspond to $g_{xt}$, $g_{ht}$. and $1 - g_{ip}$, respectively. The \textit{first} row is the input image.  The X-axis shows the ground truth output words and the Y-axis is the weight. The top-4 words with the highest scores are in red color.} \label{fig:attention}
\end{figure*}

\begin{figure}[!t]
\centering
\includegraphics[scale=0.22]{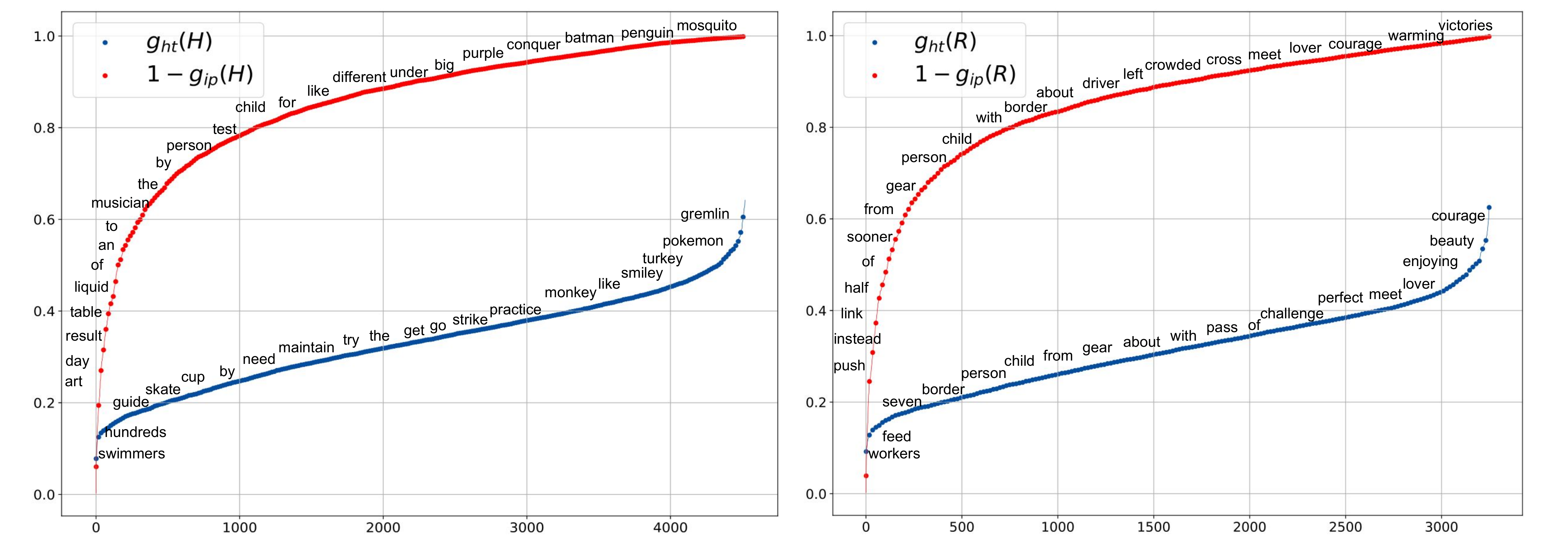}
 %where an .eps filename suffix will be assumed under latex,
 %and a .pdf suffix will be assumed for pdflatex; or what has been declared
 %via \DeclareGraphicsExtensions.

\caption{The Mean value of $1-g_{ip}$ and $g_{ht}$ for different words. \textit{Left}: humorous words. \textit{Right}: romantic words.}

%Visual features of CNN responses  and attribute detections  are injected into RNN %%%dashed arrows and get fused together through a feedback loop (blue arrows). Attention %on attributes is enforced by both input model and output model.
%}
\label{fig:meanvalue}

\end{figure}

\begin{figure}[!t]
\centering
\includegraphics[width=4.7in]{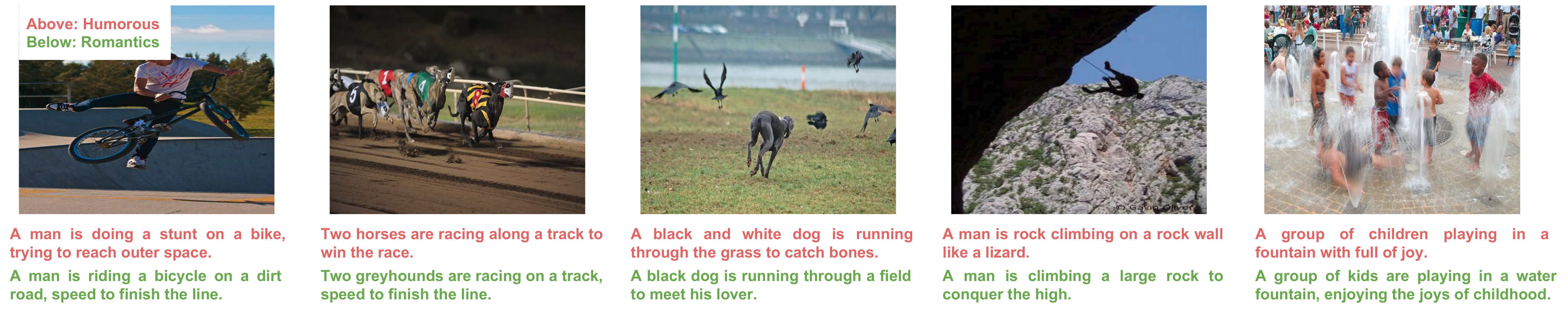}
 %where an .eps filename suffix will be assumed under latex,
 %and a .pdf suffix will be assumed for pdflatex; or what has been declared
 %via \DeclareGraphicsExtensions.

\caption{Examples of stylized captions generated by our model for different images.}

%Visual features of CNN responses  and attribute detections  are injected into RNN %%%dashed arrows and get fused together through a feedback loop (blue arrows). Attention %on attributes is enforced by both input model and output model.
%}
\label{fig_examples}

\end{figure}

In order to prove that the proposed model is effective, we visualize the attention weights of $g_{xt}$, $g_{ht}$ and $1 - g_{ip}$ mentioned in Section~\ref{sec:app} on several examples. Specifically, we directly input the ground truth stylized caption into the trained model step by step, so that at each time step, the model will give a predicted word based on the current input word and previous hidden state. This setting simulates the training process. For each time step, Figure~\ref{fig:attention} shows the ground truth output word and the corresponding attention weights. From the first example, we could see that when the model aims to predict stylized words, ``seeing'', ``their'', ``favourite'', ``player'', $g_{xt}$ (red line) and $g_{ht}$ (green line) increase remarkably, indicating that when the model predicts these words, it pays more attention to the $S_{x\cdot}$ and $S_{h\cdot}$ matrices, which capture the stylized information. Otherwise, it will focus more on $W_{x\cdot}$ and $W_{h\cdot}$, which are learned to generate factual words. On the other hand, from the fourth row, when it aims to generate words ``air'', ``when'',  ``their'', ``favourite'', the predicted word probability distribution similarity between the real and reference models is very low, this encourages the model to directly learn to generate these words by the MLE loss. Otherwise, it will pay considerable attention to the output of the reference model, which contains knowledge learned from ground truth factual captions. For the other three examples, still, when generating stylized phrases (i.e. ``looking for a me'', ``celebrating the fun of childhood'' and ``thinks ice cream help''), overall, the style-factual LSTM can effectively give more attention to $S_{x\cdot}$ and $S_{h\cdot}$, such that it will be trained mostly by corresponding ground truth words. When generating non-stylized words, the model will focus more on the factual part in the training and predicting process. It should be noticed that the first word always gets a relative high value for $g_{xt}$.  This is reasonable because it is usually the same word (i.e. ``a'') for both factual and stylized captions, the model thus cannot learn to give more attention to fact-related matrices at this very beginning. Also, some articles and prepositions, such as ``a'', ``of'', has low $1-g_{ip}$ even if they belong to a stylized phrase. This is also reasonable and acceptable, because both the real model and reference model can predict it, there is no need to pay all the attention to the corresponding ground truth stylized word.  

To further substantiate that our model successfully differentiates between stylized words and factual words, following the visualization process, we compute the mean value of $1-g_{ip}$ and $g_{ht}$ for each word in stylized dataset. As Figure~\ref{fig:meanvalue} shows, words that appear frequently in the stylized parts but rarely in the factual parts tend to get higher $g_{ht}$. Such as ``gremlin'', ``pokeman'', ``smiley'' in humorous sentences and ``courage'', ``beauty'', ``lover'' in romantic sentences. Words that appear in the stylized and factual parts with similar frequencies are likely to hold neutral value, such as ``with'', ``go'', ``of'', ``about''. Words such as ``swimmer'', ``person'', ``skate'', ``cup'', which appear mostly in the factual parts rather than the stylized parts, tend to have lower $g_{ht}$ scores. Since $g_{ht}$ represents the stylized weights in the style-factual LSTM, the result of $g_{ht}$ substantiates that the style-factual LSTM is able to differentiate between stylized and factual words. When it comes to $1-g_{ip}$, the first kind of words we mentioned above still receive high scores. However, we do not observe any clear border between the second and third kinds of words as $g_{ht}$ shows. Still, we attribute it to the fact that predicting a factual noun is overall more difficult than predicting an article or preposition, which makes its corresponding inner product lower, and thus makes $1-g_{ip}$ higher.

To make our discussion more intuitive, we show several stylized captions generated by our model in Figure~\ref{fig_examples}. As Figure~\ref{fig_examples} shows, our model can generate stylized captions that accurately describe the corresponding images. For different images, the generated captions contain appropriate humorous phrases like ``reach outer space'',``catch bones'',``like a lizard'' and appropriate romantic phrases like ``to meet his lover'',``speed to finish the line'',``conquer the high''.

\subsection{Performance on Image Sentiment Captioning Dataset}

\begin{table}[htbp]
  \small\centering
  \caption{\label{tab:result2} BLEU-1,2,3,4, ROUGE, CIDEr, METEOR scores of the proposed model and the state-of-the-art methods for sentiment captioning.}

  \begin{threeparttable}
  \scalebox{0.86}{
  \begin{tabular}{|c|c|c|c|c|c|c|c|}

    \cline{1-8}
    Model&BLEU-1&BLEU-2&BLEU-3&BLEU-4&ROUGE&CIDEr&METEOR\\ \cline{1-8}
    \multicolumn{8}{|c|}{POS Test Set} \\ \cline{1-8}
    
    NIC &48.7&28.1&17.0&10.7&36.6&55.6&15.3\\ 
    ANP-Replace &48.2&27.8&16.4&10.1&36.6&55.2&16.5\\ 
    ANP-Scoring &48.3&27.9&16.6&10.1&36.5&55.4&16.6\\ 
    LSTM-Transfer &49.3&29.5&17.9&10.9&37.2&54.1&\textbf{17.0}\\ 
    SentiCap&49.1&29.1&17.5&10.8&36.5&54.4&16.8\\ \cline{1-8}
    SF-LSTM + Adap (ours)&\textbf{50.5}&\textbf{30.8}&\textbf{19.1}&\textbf{12.1}&\textbf{38.0}&\textbf{60.0}&16.6\\ \cline{1-8}
    
\cline{1-8}
    \multicolumn{8}{|c|}{NEG Test Set} \\ \cline{1-8}
    %Model&BLEU-1&BLEU-2&BLEU-3&BLEU-4&ROUGE&CIEDr&METEOR\\ \cline{1-8}
        NIC &47.6&27.5&16.3&9.8&36.1&54.6&15.0\\ 
    ANP-Replace &48.1&28.8&17.7&10.9&36.3&56.5&16.0\\ 
    ANP-Scoring &47.9&28.7&17.7&11.1&36.2&57.1&16.0\\ 
    LSTM-Transfer &47.8&29.0&18.7&12.1&36.7&55.9&16.2\\ 
    SentiCap&50.0&\textbf{31.2}&\textbf{20.3}&13.1&37.9&\textbf{61.8}&\textbf{16.8}\\ \cline{1-8}
    SF-LSTM + Adap (ours)&\textbf{50.3}&31.0&20.1&\textbf{13.3}&\textbf{38.0}&59.7&16.2\\ \cline{1-8}
    
\cline{1-8}

    \end{tabular}}%
  \end{threeparttable}

\end{table}%

We also evaluate our model on the image sentiment caption dataset which is collected by  \cite{mathews2016senticap}. Following \cite{mathews2016senticap}, we compare the proposed model with several baselines. Besides NIC, ANP-Replace is based on NIC. For each caption generated by NIC, it randomly chooses a noun and adds the most common adjective of the corresponding sentiment for the chosen noun. In a similar way, ANP-Scoring uses multi-class logistic regression to select the most likely adjective for the chosen noun. LSTM-Transfer earns a fine-tuned LSTM from the sentiment dataset with additional regularization as \cite{schweikert2009empirical}. Senticap implements a switching LSTM with word-level regularization to generate stylized captions. It should be mentioned that Senticap utilizes ground truth word sentiment strength in their regularization, which are labeled by humans. In contrast, our model only needs ground truth image-caption pairs without extra information. 

\begin{figure}[!t]
\centering
\includegraphics[width=4.7in]{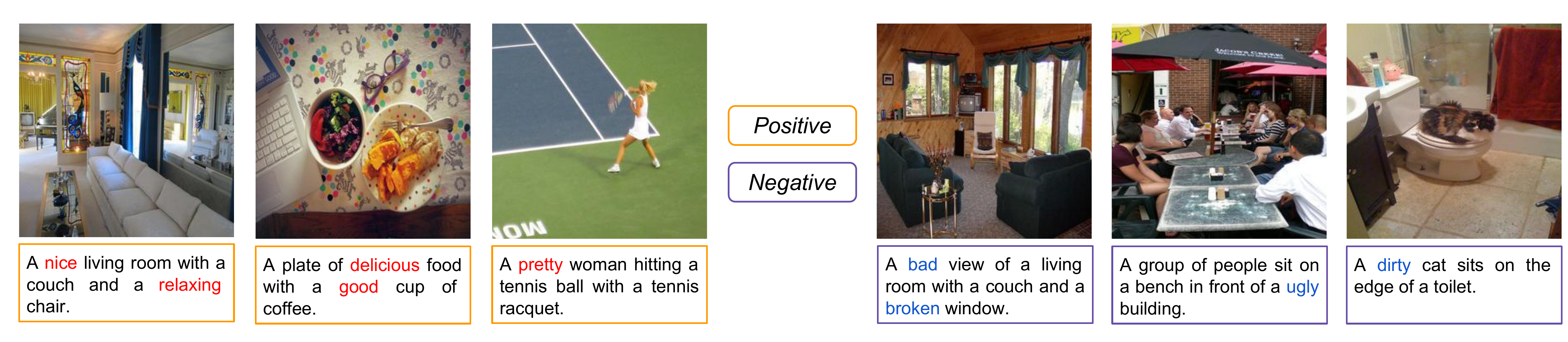}
 %where an .eps filename suffix will be assumed under latex,
 %and a .pdf suffix will be assumed for pdflatex; or what has been declared
 %via \DeclareGraphicsExtensions.

\caption{Examples of sentiment caption generation based on our model. Positive and negative words are highlighted in red and blue colors.}
\label{fig:senti_exp}

\end{figure}

Table~\ref{tab:result2} shows the performance of different models on the sentiment captioning dataset. The performance score of all baselines are directly cited from \cite{mathews2016senticap}. We can see that for positive caption generation, the performance of our proposed model remarkably outperforms other baselines, with the highest scores by almost all measures. For negative caption generation, the performance of our model is competitive with Senticap while outperforming all others. Overall, without using extra ground truth information, our model achieves the best performance for generating image captions with sentiment. Figure. \ref{fig:senti_exp} illustrates several sentiment captions generated by our model, as it can effectively generate captions with the sentiment elements being specified.

\section{Conclusions}
In this paper, we present a new stylized image captioning model. We design a style-factual LSTM as the core building block of the model, which feeds two groups of matrices into the LSTM to capture both factual and stylized information. To allow the model to preserve factual information in a better way, we leverage the reference model and develop an adaptive learning approach to adaptively adding factual information into the model, based on the prediction similarity between the real and reference models. Experiments on two stylized image captioning datasets demonstrate the effectiveness of our proposed approach. It outperforms the state-of-the-art models for stylized image captioning without using extra ground truth information. Furthermore, visualization of different attention weights demonstrates that our model can indeed differentiate the factual part and stylized part of a caption automatically, and adjust the attention weights adaptively for better learning and prediction.     

\section{Acknowledgment}
We would like to thank the support of New York State through the Goergen Institute for Data Science, our corporate sponsor Adobe and NSF Award \#1704309.

\clearpage

\bibliographystyle{splncs04}
\bibliography{egbib}
\end{document}